\documentclass[11pt,journal,draftcls,letterpaper,onecolumn]{IEEEtran}
\ifCLASSINFOpdf
\else
\fi
\usepackage{amsmath}
\usepackage{amsfonts}
\usepackage{amssymb}
\usepackage{stmaryrd}

\usepackage{algorithmic}
\usepackage{url}
\usepackage{graphicx}

\begin{document}
%
\title{UCB Algorithm for Exponential Distributions}
%
%
%

\author{Wassim~Jouini
        and~Christophe Moy
\thanks{SUPELEC, IETR, SCEE, Avenue de la Boulaie, CS 47601, 35576
Cesson S\'evign\'e, France.}
\thanks{Email: {wassim.jouini}@supelec.fr}
}
\maketitle

\begin{abstract}
We introduce in this paper a new algorithm for Multi-Armed Bandit (MAB) problems.  A machine learning paradigm popular within Cognitive Network related topics (e.g., Spectrum Sensing and Allocation).  We focus on the case where the rewards are exponentially distributed, which is common when dealing with Rayleigh fading channels. This strategy, named Multiplicative Upper Confidence Bound (MUCB), associates a utility index to every available arm, and then selects the arm with the highest index. For every arm, the associated index is equal to the product of a multiplicative factor by the sample mean of the rewards collected by this arm. We show that the MUCB policy has a low complexity and is order optimal.
\end{abstract}


\IEEEpeerreviewmaketitle

\section{Introduction}
\newcommand{\E}{\mathbb{E}}
\renewcommand{\P}{\mathbb{P}}
\newcommand{\1}{{\rm 1\hspace*{-0.4ex}
\rule{0.1ex}{1.52ex}\hspace*{0.2ex}}}
\newtheorem{lemma}{Lemma}
\newtheorem{theorem}{Theorem}
\newtheorem{assumption}{Assumption}
\newtheorem{fact}{Fact}  
\newtheorem{definition}{Definition}

\par Several sequential decision making problems face a dilemma between the
exploration of a space of  choices, or solutions, and the exploitation
of  the information  available  to the  decision  maker.  The  problem
described  herein  is  known   as  sequential  decision  making  under
uncertainty. In  this paper we focus  on a sub-class  of this problem,
where the decision  maker has a discrete set  of stateless choices and
the  added information  is a  real valued  sequence (of  feedbacks, or
rewards) that  quantifies how well  the decision maker behaved  in the
previous time  steps. This particular instance  of sequential decision
making  problems is generally  known as  the multi-armed  bandit (MAB)
problem \cite{Robbins1985,Agrawal1995}.   
\par A   common  approach  to  solving  the  exploration  versus
exploitation dilemma within MAB problems consists in assigning an utility  value to every
arm. An  arm's utility aggregates  all the past information  about the
lever  and quantifies  the  gambler's interest  in  pulling it.   Such
utilities  are called \emph{indexes}.  Agrawal et al. \cite{Agrawal1995} emphasized
the  family of  indexes  minimizing the  expected  cumulated loss  and
called them Upper Confidence Bound (UCB) indexes.  UCB indexes provide
an optimistic  estimation of the  arms' performances while  ensuring a
rapidly  decreasing probability  of selecting  a suboptimal  arm.  The
decision  maker builds its  policy by  greedily selecting  the largest
index.  Recently, Auer et al. \cite{Auer2002} proved that a simple additive form, 
 of the rewards' sample mean and a bias, known as $UCB_1$ can achieve order optimality over time when 
 dealing with rewards drawn from bounded distributions.  Tackling exponentially distributed 
 rewards remains however a challenge as optimal learning algorithms to tackle this matter prove to be complex to implement \cite{Robbins1985,Agrawal1995}.

\par This  paper  is  inspired  from  the
aforementioned  work.    However,  we   suggest  the  analysis   of  a
multiplicative rather than an additive  expression for the index. 
\par  The main contribution  of this  paper is to  design and analyze  a simple,
deterministic, multiplicative index-based  policy. The decision making
strategy computes an index associated to every available arm, and then
selects the arm  with the highest index. Every  index associated to an
arm is equal to the product of the sample mean of the reward collected
by this arm  and a scaling factor. The scaling factor  is chosen so as
to  provide   an  optimistic   estimation  of  the   considered  arm's
performance.   
\par We  show  that  our  decision policy  has  both  a  low
computational complexity and can lead  to a logarithmic loss over time
under some non-restrictive  conditions. For the rest  of this paper
we will  refer to our  suggested policy as  \emph{Multiplicative Upper
Confidence Bound index} (MUCB).

\par The outline  of this  paper is the  following: We start  by presenting
some  general   notions  on   the  multi-armed  bandit   framework with exponentially distributed rewards in
Section~\ref{sec:mab}.  Then,  Section~\ref{sec:mucb}  introduces  our
index  policy and  Section~\ref{sec:analysis} analyzes  its behavior,
proving the order optimality of the suggested algorithm. Finally,
Section~\ref{sec:conc}concludes.

\section{Multi-Armed Bandits}
\label{sec:mab}
A $K$-armed bandit ($K \in  \mathbb{N}$) is a machine learning problem
based  on an  analogy  with the  traditional  slot machine  (one-armed
bandit) but with more than one lever. Such a problem is defined by the
$K$-tuple  $(\theta_{1},  \theta_{2},...,  \theta_{K}) \in  \Theta^K$,
$\Theta$  being the  set of  all positive  reward  distributions. When
pulled  at  a time  $t  \in  \mathbb{N}$,  each lever\footnote{We  use
  indifferently the  words ``lever'',  ``arm'', or ``machine''.}  $k \in \llbracket  1, K  \rrbracket $
(where $\llbracket 1 , K  \rrbracket = \{1,...,K\}$) provides a reward
$r_{t}$  drawn from  a  distribution $\theta_{k}$  associated to  that
specific  lever. The  objective  of  the gambler  is  to maximize  the
cumulated  sum of rewards  through iterative  pulls.  It  is generally
assumed that the  gambler has no (or partial)  initial knowledge about
the levers.  The  crucial tradeoff the gambler faces  at each trial is
between \emph{exploitation} of the lever that has the highest expected
payoff  and  \emph{exploration}  to  get more  information  about  the
expected payoffs of  the other levers.  In this  paper, we assume that
the  different  exponentially distributed payoffs  drawn  from  a machine  are  independent  and
identically  distributed (i.i.d.)   and that  the independence  of the
rewards holds  between the  machines. However the  different machines'
reward  distributions $(\theta_{1},  \theta_{2},...,  \theta_{K})$ are
not supposed to be the same.

Let $I_t \in \llbracket 1, K\rrbracket$ denote the machine selected at
a time  $t$, and let  $H_t$ be the  {history} vector available  to the
gambler at instant $t$, i.e., $H_t=[I_0,r_{0},I_1,r_{1},\ldots,I_{t-1},r_{t-1}]$

\par We assume that the gambler uses  a policy $\pi$ to select arm $I_t$ at
instant $t$, such that $I_t=\pi(H_t)$.  We shall also write $\forall k
\in  \llbracket 1,  K\rrbracket, \  \mu_k {\buildrel  \Delta  \over =}\frac{1}{\lambda_k}{\buildrel  \Delta  \over =}
\E[\theta_{k}]$, where $\lambda_k$ refers to the parameter of the considered exponential distribution with pdf $f_{\theta_k}(x)=\lambda_k e^{-\lambda_{k}{x}}, \ x\geq0$,  and  we  assume  that  $\mu_k  >  0,  \forall  k  \in
\llbracket 1, K \rrbracket .$ The (cumulated) regret of a policy $\pi$
at time $t$ (after $t$ pulls) is defined as follows: $R_t=t\mu^{*}-\sum^{t-1}_{m=0}r_{m}$,
where  $\mu^*= \underset {k \in \llbracket  1, K \rrbracket}
{\max} \left\{  \mu_{k}\right\}$ refers to the expected  reward of the
optimal arm.

\par We seek to find a  policy that minimizes the expected cumulated regret
(Equation~\ref{equ_Exp_regret}),
\begin{eqnarray}
\E\left[ R_t \right] = \sum_{k\neq k^{*}}\Delta_{k}\E\left[T_{k,t}\right],
\label{equ_Exp_regret}  
\end{eqnarray}
where  $\Delta_{k}=\mu^*-\mu_k$ is  the expected  loss of  playing arm
$k$, and $T_{k,t}$  refers to the number of  times the machine $k$
has been  played from  instant $0$ to  instant $t-1$.

\section{Multiplicative upper confidence bound algorithms}
\label{sec:mucb}
This section presents our main contribution, the introduction of a new
multiplicative index.  Let $B_{k,t}(T_{k,t})$  denote the index of arm
$k$  at  time  $t$ after  being  pulled  $T_{k,t}$.   We refer  to  as
Multiplicative Upper Confidence Bound  algorithms (MUCB) the family of
indexes that can be written in the form:
\begin{equation*}
B_{k,t}\left(T_{k,t}\right)=\overline{X}_{k,t}(T_{k,t})M_{k,t}\left(T_{k,t}\right),
\end{equation*}
where $\overline{X}_{k,t}(T_{k,t})$ is the  sample mean of machine $k$
at step  $t$ after $T_{k,t}$  pulls, i.e.,
$\overline{X}_{k,t}(T_{k,t}) = \frac {1} {T_{k,t}} \sum_{i=0}^{t-1}
\1_{\{I_i = k\}} r_i$ and   $M_{k,t}(\cdot)$  is   an  upper   confidence   scaling  factor
chosen to  insure that the  index $B_{k,t}(T_{k,t})$ is  an increasing
function of the number of  rounds $t$. This last property insures that
the index  of an arm  that has  not been pulled  for a long  time will
increase, thus  eventually leading  to the sampling  of this  arm.  We
introduce a particular parametric class of MUCB indexes, which we call
$MUCB(\alpha)$, given as follows\footnote{This form offers a compact mathematical formula. However practically speaking, a machine $k$ is played when $T_{k,t}\leq\alpha\ln(t)$. Otherwise the machine with largest finite index is played.}:
\begin{equation}
\forall \alpha \geq 0, \
M_{k,t}\left(T_{k,t}\right)=\frac{1}{\max
    \left\{0;(1-\sqrt{\frac{\alpha\ln(t)}{T_{k,t}}})\right\}}
\ 
\label{equ_index_sucb_2}
\end{equation}
We  adopt   the  convention  that   $\frac{1}{0}  =
+\infty$.  
Given a  history $H_t$,  one can compute  the values of  $T_{k,t}$ and
$M_{k,t}$ and derive an index-based policy $\pi$ as follows:
\begin{equation}
I_t = \pi(H_t) \in \underset {k \in \llbracket 1, K \rrbracket} {\arg
  \max}\left\{B_{k,t}\left(T_{k,t}\right)\right\}.
\label{equ_policy2}
\end{equation}

\section{Analysis of $MUCB(\alpha)$ policies}
\label{sec:analysis}
This section analyses the theoretical properties of $MUCB(\alpha)$
algorithms. More  specifically, it focuses on determining  how fast is
the  optimal  arm  identified   and  what  are  the  probabilities  of
\emph{anomalies}, that is sub-optimal pulls. 
\subsection{Consistency and order optimality of MUCB indexes}
\label{sec:consistency}

\begin{definition}[$\beta$-consistency]Consider the  set $\Theta^K$ of
  $K$-armed  bandit   problems.   A  policy   $\pi$  is  said   to  be
  \textit{$\beta$-consistent}, $  0 < \beta  \leq 1$, with  respect to
  $\Theta^K$, if and only if
  \begin{equation} \forall (\theta_1,\ldots, \theta_K) \in \Theta^K,
    \lim_{t \rightarrow \infty} \frac{\E[R_{t}]}{t^\beta}=0
\label{equ_consistent}
\end{equation}
\end{definition}
\vspace{0.1cm}
We expect  good policies  to be at  least \textit{1-consistent}.  As a
matter of  fact, \textit{1-consistency} ensures  that, asymptotically,
the average expected reward is optimal.
\vspace{0.1cm}
\par From the  expression of  Equation \ref{equ_Exp_regret} one  can remark
that its  is sufficient  to upper bound  the expected number  of times
$\E[T_{k,t}]$ one plays a suboptimal  machine $k$ after $t$ rounds, to
obtain an upper bound on  the expected cumulated regret. This leads to
the following theorem.

\begin{theorem}[Order optimality of $MUCB(\alpha)$ policies]
Let $\rho_k=\mu_{k}/\mu^{*}$, $k \in \llbracket 1, K \rrbracket \setminus
  \{k^*\}$. For all $K \geq 2 $, if policy \hbox{$MUCB(\alpha>4)$} is run on $K$ machines having rewards drawn from exponential
distributions $\theta_1, . . . , \theta_K$ then:
\begin{align}
	\E\left[ R_t \right]\leq \sum_{k: \Delta_{k}>0}\frac{4\mu^{*}\alpha}{1-\rho_{k}}\ln(t)+o\left(\ln(t)\right)
	\label{equ_regret_ucb_1}
\end{align}
\label{thm_MUCB} 
\end{theorem}

\vspace{0.1cm}
Proving  Theorem \ref{thm_MUCB} relies on three lemmas that we analyze and prove in the next subsection. The lemma \ref{lemma:upperbound} provides a general bound for the regret regardless of the policy considered. The expression is function of two probabilities related to learning anomalies. These anomalies depend on the learning algorithm. They are introduced and analyzed. Then through lemma \ref{lemma:anomaly1} ad \ref{lemma:anomaly2} we upper bound them. 

\subsection{Learning Anomalies and Consistency of MUCB policies}
\label{subsec:anomalies}
Let  us  introduce  the  set  $\mathbb{S}=\mathbb{N}\times\mathbb{R}$;
then, one  can write  $S_{k,t}=(T_{k,t}, B_{k,t}) \in  \mathbb{S}$ the
decision state of arm $k$ at time $t$.
We associate the product order to  the set $\mathbb{S}$: for a pair of
states $S =  (T,B) \in \mathbb{S}$ and $S'  = (T',B') \in \mathbb{S}$,
we write $S\geq S'$ if and only if $T \geq T'$ and $B \geq B'$.

\begin{definition}[Anomaly of type 1]
  We assume that  there exists at least one  suboptimal machine, i.e.,
  $\llbracket 1,  K \rrbracket \setminus \{k^*\}  \neq \emptyset$.  We
  call {\it anomaly of  type 1}, denoted by $\{\phi_{1}(u_k)\}^{\pi}_{k,t}$,
  for a suboptimal machine $k \in \llbracket 1, K \rrbracket \setminus
  \{k^*\}$,  and with  parameter $u_k  \in \mathbb{N}$,  the following
  event:
\begin{eqnarray}
\left\{\phi_{1}\left(u_k\right)\right\}^{\pi}_{k,t}=\left\{S_{k,t} \geq
  (u_k, \mu^*) \right\} \ .  \nonumber
\end{eqnarray}
\end{definition}


\begin{definition}[Anomaly of type 2]
  We   refer   to  as   {\it   anomaly   of   type  2},   denoted   by
  $\{\phi_{2}\}^{\pi}_t$, associated to the optimal machine $k^*$, the
  following event:
\begin{equation*}
\left\{\phi_{2}\right\}^{\pi}_t=\left\{S_{k^*,t} < (\infty , \mu^*)
  \ \cap \ T_{k^*,t} \geq 1 \right\} \ .
\end{equation*}
\end{definition}
\vspace{0.1cm}

\begin{lemma}[Expected cumulated regret. Proof in \ref{subsec:Lemma1}]
\label{lemma:upperbound}
  Given a policy $\pi$ and a  MAB problem, let ${\bf u} = [u_1,\ldots,u_K]$ represent a set of integers, then the  expected cumulated
  regret  is upper  bounded by:
\begin{equation*}
  \E[R_t ] \leq \sum_{k\neq k^{*}} \Delta_{k}u_k
  + \sum_{k\neq k^{*}} \Delta_{k} \P_t(u_k) 
	\label{regret_complet}
\end{equation*}
with, $\P_t(u_k)=\sum^{t}_{m=u_k+1} \left( \P\left(\{
      \phi_{2} \}^{\pi}_m\right)
    + \P\left( \{\phi_{1}(u_k)\}^{\pi}_{k,m} \right)  \right)$
\end{lemma}

We consider the following values for the set $\bf{u}$, for all suboptimal arms $k$, 
$
u_k(t)=\left\lceil \frac{4\alpha}{\left(1-\rho_{k}\right)^2}\ln(t)\right\rceil
$.

We show in the two following lemmas that for the defined set $\bf{u}$ the anomalies are upper bounded by exponentially decreasing functions of the number of iterations.

\begin{lemma}[Upper bound of Anomaly 1. Proof in \ref{subsec:Lemma2}]
For all $K \geq 2 $, if policy \hbox{$MUCB(\alpha)$} is run on $K$ machines having rewards drawn from exponential
distributions $\theta_2, . . . , \theta_K$ then $\forall k \in \llbracket 1, K \rrbracket \setminus
  \{k^*\}$:
\begin{align}
\P\left( \{\phi_{1}(u_k)\}^{\pi}_{k,t} \right) \leq t^{-\alpha/2+1}	
\end{align}
\label{lemma:anomaly1}
\end{lemma}
\begin{lemma}[Upper bound of Anomaly 2. Proof in \ref{subsec:Lemma3}]
For all $K \geq 2 $, if policy \hbox{$MUCB(\alpha)$} is run on $K$ machines having rewards drawn from exponential
distributions $\theta_1, . . . , \theta_K$ then:
\begin{align}
	\P\left(\{\phi_{2} \}^{\pi}_t\right) \leq t^{-\alpha/2+1}	
\end{align}
\label{lemma:anomaly2}
\end{lemma}
\par We end this paper by the proof of Theorem \ref{thm_MUCB}.
\begin{proof}[Proof of Theorem \ref{thm_MUCB}]
For $\alpha>4$, relying on Lemmas \ref{lemma:upperbound}, \ref{lemma:anomaly1} and \ref{lemma:anomaly2} we can write: 
$$
  \E[R_t ] \leq \sum_{k\neq k^{*}} \Delta_{k}\left\lceil \frac{4\alpha}{\left(1-\rho_{k}\right)^2}\ln(t)\right\rceil + o(\ln(t)) 
$$
with, $\sum_{k\neq k^{*}} \Delta_{k}\P_t(u_k)=o(\ln(t))$. Finally, since $\Delta_k=\mu^{*}(1-\rho_k)$ and $u_k(t)=\frac{4\alpha}{\left(1-\rho_{k}\right)^2}\ln(t)+o(ln(t))$, we find the stated result in Theorem \ref{thm_MUCB}.
\end{proof}

\section{Conclusion}
\label{sec:conc}
A new low complexity algorithm for MAB problems is suggested and analyzed in this paper: MUCB. The analysis of its regret proves that the algorithm is order optimality over time. In order to quantify it performance compared to optimal algorithms, further empirical evaluations are needed and are currently under investigation. 
\section*{Acknowledgment}

The authors would like to thank Damien Ernst, Raphael Fonteneau and Emmanuel Rachelson for their many helpful comments and answers regarding this work.

\bibliographystyle{unsrt}

\bibliography{cognitive}

\begin{thebibliography}{1}

\bibitem{Robbins1985}
T.L. Lai and H.~Robbins.
\newblock Asymptotically efficient adaptive allocation rules.
\newblock {\em Advances in Applied Mathematics}, 6:4--22, 1985.

\bibitem{Agrawal1995}
R.~Agrawal.
\newblock {Sample mean based index policies with O(log(n)) regret for the
  multi-armed bandit problem}.
\newblock {\em Advances in Applied Probability}, 27:1054--1078, 1995.

\bibitem{Auer2002}
P.~Auer, N.~Cesa-Bianchi, and P.~Fischer.
\newblock Finite time analysis of multi-armed bandit problems.
\newblock {\em Machine learning}, 47(2/3):235--256, 2002.

\bibitem{Chernoff1952}
H.~Chernoff.
\newblock A measure of asymptotic efficiency fo tests of a hypothesis based on
  the sum of observations.
\newblock {\em The Annals of Mathematical Statistics}, pages 493--507, 1952.

\end{thebibliography}

\section{Appendix}
\subsection{Large deviations inequalities}
\label{subsec:large_deviations_inequalities}

\begin{assumption}[Cramer condition] \label{assumption:cramer} Let $X$
  be a real random variable. $X$ satisfies the Cramer condition if and
  only if
\begin{eqnarray}
\exists \gamma > 0 :  \forall \eta \in (0,\gamma),   \mathbb E
 \left[   e^{\eta X} \right] < \infty \ .  \nonumber
\end{eqnarray}
\end{assumption}

\begin{lemma}[Cramer-Chernoff   Lemma     for     the     sample
  mean]
\label{lemma:chernoff}  Let  $X_1,  \ldots,  X_n$  $(n  \in
  \mathbb N)$ be a sequence of i.i.d. real random variables satisfying
  the Cramer  condition with expected value $\mathbb  E[X]$. We denote
  by $\overline{X}_n$ the sample  mean $\overline{X}_n = \frac {1} {n}
  \sum_{i=1}^{n} X_i$.   Then, there exist  two functions $l_1(\cdot)$
  and $l_2(\cdot)$ such that:
 \begin{eqnarray}
 \forall \beta_1 > \E[X],  \P(\overline{X}_{n}  \geq   \beta_1)\leq
 e^{-l_1(\beta_1) n} \ , \nonumber 
 \end{eqnarray}
\begin{eqnarray}
  \forall \beta_2 < \E[X], \P(\overline{X}_{n}    \leq    \beta_2)\leq
  e^{-l_2(\beta_2) n} \ . \nonumber 
\end{eqnarray}
Functions $l_1(\cdot)$  and $l_2(\cdot)$ do  not depend on  the sample
size $n$ and are continuous non-negative, strictly increasing (respectively strictly-decreasing) for all $\beta_1>\E(X)$ (respectively $\beta_2<\E(X)$), both null for $\beta_1=\beta_2=\E(X)$.
\par This result was initially  proposed and proved in \cite{Chernoff1952}.
The bounds  provided by  this lemma are  called {\it  Large Deviations
Inequalities} (LDIs) in this paper.
\end{lemma}
\par In the case of exponential distributions this theorem can be applied and LDI functions have the following expressions:
$$l_1(\beta)=l_2(\beta)= \frac{\beta}{\E[X]}-1-\ln\left(\frac{\beta}{\E[X]}\right)\geq \frac{3\left(1-\frac{\beta}{\E[X]}\right)^2}{2\left(1+2\frac{\beta}{\E[X]}\right)}$$
\subsection{Proof of Lemma \ref{lemma:upperbound}}
\label{subsec:Lemma1}
According  to  Equation  \ref{equ_Exp_regret}:
$
\E[R^{\pi}_t]  =  \sum\limits_{k\neq k^{*}}
  \Delta_k \E \left[ T_{k,t}\right] \ 
$.
Per definition $T_{k,t}  = \sum\limits_{m=0}^{t-1} \1_{I_m = k}$.  Then,
$
\E[T_{k,t}] = \sum\limits_{m=0}^{t-1}\E \left[ \1_{I_m = k}\right]
$.
After playing an arm $u_k$ times,  bounding the
first $u_k$ terms by 1 yields:

\begin{equation}
\E[T_{k,t}] \leq u_k +  \sum\limits_{m=u_k+1}^{t-1} \P \left( \left\{I_m
  = k \right\} \cap \left\{T_{k,m}>u_k\right\}\right)
\label{eq:ET_k}
\end{equation}

Then we can notice that the following events are equivalent:
$$
\left\{I_m = k\right\}=\left\{B_{k,m} > \max\limits_{k'\neq k} B_{k',m}\right\}
$$
Moreover we can notice that:
$$
\left\{B_{k,m} > \max\limits_{k'\neq k} B_{k',m}\right\}\subset
\left\{B_{k,m} >  B_{k*,m}\right\}
$$
Which can be further included in the following union of events:
$$
\left\{B_{k,m} >  B_{k*,m}\right\}\subset \left\{B_{k,m} > \mu^*\right\} \cup \left\{\mu^* >  B_{k*,m}\right\}
$$
Consequently we can write:
\begin{align}
\left\{I_m  = k \right\} \cap \left\{T_{k,m}>u_k\right\} \subset
\left\{\Phi_1(u_k) \right\}^{\pi}_{k,m} \cup \left\{\Phi_2
  \right\}^{\pi}_{m}
\end{align}

Finally, we apply the probability operator: 
\begin{eqnarray}
\E[T_{k,t}] \leq u_k +  \sum\limits_{m=u_k+1}^{t-1}
\P(\left\{ \Phi_1(u_k)  \right\}^{\pi}_{k,m})  +
\P(\left\{ \Phi_2 \right\}^{\pi}_m ) \ . \label{thm:bound_regret_04}
\end{eqnarray}
The  combination   of  Equation  \ref{equ_Exp_regret} -  given  at  the
beginning  of this proof -  and Equation  \ref{thm:bound_regret_04}
concludes this proof.

\subsection{Proof of Lemma \ref{lemma:anomaly1}}
\label{subsec:Lemma2}
From the definition of $\{\phi_{1}(u_k)\}^{\pi}_{k,t}$ we can write that: 
\begin{align*}
\P\left(\{\phi_{1}(u_k)\}^{\pi}_{k,t}\right) &= \sum_{S_{k,t} \in
  \mathbb{S}} \P\left(S_{k,t} \geq \left(u_k,\mu^*\right)\right), \\ 
& \leq \sum^{t-1}_{u=u_k} \P\left(B_{k,t}(u)\geq \mu^* \right).
\end{align*}
In the case of MUCB policies, we have:
\begin{equation*}
\forall u\leq t, \  \P\left(B_{k,t}(u)\geq
  \mu^*\right)=\P\left(\overline{X}_{k,t}(u) \geq
  \frac{\mu^{*}}{M_{k,t}(u)}.\right) 
\end{equation*}
Consequently, we can upper bound the probability of occurrence of type
1 anomalies by: 
\begin{equation*}
\P\left(\{\phi_{1}(u_k)\}^{\pi}_{k,t}\right) \leq
\sum^{t-1}_{u=u_k}\P\left(\overline{X}_{k,t}(u) \geq
  \frac{\mu^{*}}{M_{k,t}(u)}\right). 
\end{equation*}
Let us define $\beta_{k,t}(T_{k,t})=\frac{\mu^{*}}{M_{k,t}(T_{k,t})}$.

Since we are dealing with exponential distributions,  the rewards provided  by the arm $k$ satisfy the Cramer
condition.  As a matter of fact, since $u\geq u_k \geq \alpha\frac{\ln(t)}{\left(1-\rho_{k}\right)^2}$ then: 
$$\beta_{k,t}(u)\lambda_k=\rho_{k}^{-1}\left(1-\sqrt{\alpha\frac{\ln(t)}{u}}\right)\geq 1$$
So,  according  to  the large  deviation  inequality  for
$\overline{X}_{k,t}(T_{k,t})$  given   by  Lemma  \ref{lemma:chernoff}
(with  $T_{k,t} \geq  u_k$ and  $u_k$  large enough),  there exists  a
continuous, non-decreasing, non-negative function $l_{1,k}$ such that:
\begin{equation*}
\P\left(\overline{X}_{k,t}(T_{k,t})\geq \beta_{k,t}(T_{k,t}) |
  T_{k,t}=u \right) \leq e^{-{l_{1,k}(\beta_{k,t}(u))}u}.
\end{equation*}
Finally: 
\begin{equation}
 \P\left(\{\phi_{1}(u_k)\}^{\pi}_{k,t}\right)\leq \sum^{t-1}_{u=u_k}
 e^{-{l_{1,k}(\beta_{k,t}(u))}u}. 
\label{Eq:FundInegality}
\end{equation}

The end of this proof aims at proving that for $u\geq u_k$: $l_{1,k}(\beta_{k,t}(u))\geq\alpha\frac{\ln(t)}{2u}$.
\par Note that since we are dealing with exponential distributions we can write: $l_{1,k}(\beta_{k,t}(u))\geq \frac{3\left(1-\beta_{k,t}(u)\lambda_{k}\right)^{2}}{2\left(1+2\beta_{k,t}(u)\lambda_{k}\right)}$.

Moreover since $u\geq u_k \geq \alpha\frac{\ln(t)}{\left(1-\rho_{k}\right)^2}$ then: 
$$\beta_{k,t}(u)\lambda_k=\rho_{k}^{-1}\left(1-\sqrt{\alpha\frac{\ln(t)}{u}}\right)\leq \rho_{k}^{-1}$$
Consequently it is sufficient to prove that:
$$
\frac{3\left(1-\beta_{k,t}(u)\lambda_{k}\right)^{2}}{2\left(1+2\rho_{k}^{-1}\right)} \geq \alpha\frac{\ln(t)}{2u}
$$
Let us define $h(t)$ as a function of time: $h(t)=\sqrt{\alpha\frac{\ln(t)}{u}} \in [0, \ 1]$.
We analyze the sign of the function: 
\begin{align}
g(t)=\left(\rho_k^{-1}h(t)-\left(\rho_{k}^{-1}-1\right)\right)^{2}-\frac{\left(1+2\rho_k^{-1}\right)}{3}h(t)^2
\end{align}
Consequently we need to prove that for $u\geq u_k$, $g(\cdot)$ has positive values.

Factorizing last equation leads to the following to terms:
\begin{equation}
\left\{
	\begin{array}{ll} 
	    \left(\rho_k^{-1}-\sqrt{\frac{\left(1+2\rho_k^{-1}\right)}{3}}\right)h(t)-\left(\rho_{k}^{-1}-1\right)\\
	    \left(\rho_k^{-1}+\sqrt{\frac{\left(1+2\rho_k^{-1}\right)}{3}}\right)h(t)-\left(\rho_{k}^{-1}-1\right)
	\end{array}
\right.
\label{Eq:EtudeFonction}
\end{equation}

Since per definition: $h(t) \in [0, 1]$ and $\rho_{k}^{-1}\geq1$ then,
$
\left(\rho_k^{-1}-\sqrt{\frac{\left(1+2\rho_k^{-1}\right)}{3}}\right)h(t)-\left(\rho_{k}^{-1}-1\right)\leq 0
$.
Consequently, $g(\cdot)$ is positive only if the second term of Equation \ref{Eq:EtudeFonction} is negative, i.e., 
$
\sqrt{\alpha\frac{\ln(t)}{u}}\leq \frac{\left(\rho_{k}^{-1}-1\right)}{\left(\rho_k^{-1}+\sqrt{\frac{\left(1+2\rho_k^{-1}\right)}{3}}\right)}
$.
Since $u\geq u_k$, the last inequation is verified.
Finally upper bounding Equation \ref{Eq:FundInegality} for $u\geq u_k$:
$$
\P\left(\{\phi_{1}(u_k)\}^{\pi}_{k,t}\right)\leq \sum^{t-1}_{u=u_k}
 e^{-\alpha\ln(u)/2}\leq \sum^{t-1}_{u=u_k}\frac{1}{u^{\alpha/2}} \leq \frac{1}{t^{\alpha/2-1}}
$$

\subsection{Proof of Lemma \ref{lemma:anomaly2}}
\label{subsec:Lemma3}
This proof follows the same steps as the the proof in Subsection \ref{subsec:Lemma2}.

From the definition of $\{\phi_{1}(u_k)\}^{\pi}_{k,t}$ we can write that:
$
\P\left(\{\phi_{2}\}^{\pi}_{t}\right) \leq \sum^{t-1}_{u=1} \P\left(B_{k^{*},t}(u)\leq \mu^* \right)
$
\par In the case of MUCB policies, we have:
\begin{equation*}
\forall u\leq t, \  \P\left(B_{k^{*},t}(u)\leq
  \mu^*\right)=\P\left(\overline{X}_{k^{*},t}(u) \leq
  \frac{\mu^{*}}{M_{k^{*},t}(u)}.\right) 
\end{equation*}
Consequently, we can upper bound the probability of occurrence of type
2 anomalies by: 
\begin{equation*}
\P\left(\{\phi_{2}\}^{\pi}_{t}\right) \leq
\sum^{t-1}_{u=1}\P\left(\frac{\overline{X}_{k^{*},t}(u)}{\mu^{*}} \leq \max\left\{0;(1-\sqrt{\frac{\alpha\ln(t)}{T_{k,t}}})\right\}
  \right)
\end{equation*}

Since $\mu^{*}\max\left\{0;(1-\sqrt{\frac{\alpha\ln(t)}{T_{k,t}}})\right\}\leq\mu^{*}$ Cramer's condition is verified.
Moreover since the machine is played when the maximal of the previous term is equal to $0$, we can consider that $u\geq\alpha\ln(t)$ and that:
$$ \mu^{*}\max\left\{0;(1-\sqrt{\frac{\alpha\ln(t)}{T_{k,t}}})\right\}= \mu^{*}\left(1-\sqrt{\frac{\alpha\ln(t)}{T_{k,t}}}\right)$$

\vspace{0.1cm}

Consequently, we can upper-bound the occurrence of Anomaly 2:
\begin{align}
\P\left(\{\phi_{2}\}^{\pi}_{t}\right) \leq \sum^{t-1}_{u=\alpha\ln(t)}
 e^{-{l_{2}(\beta_{k^{*},t}(u))}u}
\label{Eq:UBA2}
\end{align}
Where, $l_{2}(\beta_{k^{*},t}(u))$ verifies the LDI as defined in Appendix \ref{subsec:large_deviations_inequalities}. Thus, after mild simplifications we can write, 
$$
l_{2}(\beta_{k^{*},t}(u))\geq \frac{\frac{3\alpha\ln(t)}{u}}{2\left(1+2(1-\sqrt{\frac{\alpha\ln(t)}{u}})\right)}\geq \frac{\alpha\ln(t)}{2u}
$$

\vspace{0.1cm}
Consequently, including this last inequality into Equation \ref{Eq:UBA2} ends the proof.

\end{document}